\documentclass{article}
\usepackage{ijcai22}

\usepackage{times}
\usepackage{soul}
\usepackage{url}
\usepackage[hidelinks]{hyperref}
\usepackage[utf8]{inputenc}
\usepackage{graphicx}

\usepackage{amsmath}
\usepackage{amsthm}
\usepackage{booktabs}
\usepackage{algorithm}
\usepackage{algorithmic}
\usepackage{microtype}
\usepackage{natbib}
\usepackage{danudefs}
\usepackage{booktabs}
\usepackage{xspace}
\usepackage{enumitem}

\DeclareMathOperator*{\minimise}{minimise}

\usepackage[english]{babel}
\usepackage{hyperref}
\addto\captionsenglish{%
}
\addto\extrasenglish{%
}

\title{A Survey on Word Meta-Embedding Learning}

\author{
Danushka Bollegala$^{1,2}$
\and
James O'Neill$^1$
\affiliations
$^1$University of Liverpool, $^2$Amazon\\
\emails
\texttt{\{danushka, james.o-neill\}@liverpool.ac.uk}
}

\begin{document}
\maketitle
\begin{abstract}
  Meta-embedding (ME) learning is an emerging approach that attempts to learn more accurate word embeddings given existing (source) word embeddings as the sole input.
  Due to their ability to incorporate semantics from multiple source embeddings in a compact manner with superior performance, ME learning has gained popularity among practitioners in NLP.
  To the best of our knowledge, there exist no prior systematic survey on ME learning and this paper
 attempts to fill this need.
  We classify ME learning methods according to multiple factors such as whether they 
  (a) operate on static or contextualised embeddings, (b) trained in an unsupervised manner or (c) fine-tuned for a particular task/domain.
  Moreover, we discuss the limitations of existing ME learning methods and highlight potential future research directions.
\end{abstract}

\section{Introduction}
\label{sec:intro}

Given independently trained multiple word representations (aka \emph{embeddings}) learnt using diverse algorithms and lexical resources, word meta-embedding (ME) learning methods~\citep{Yin:ACL:2016,Bao:COLING:2018,Bollegala:IJCAI:2018,wu2020task,He:2020,jawanpuria-etal-2020-learning,Coates:NAACL:2018} attempt to learn more accurate and wide-coverage word embeddings. 
The input and output word embeddings to the ME algorithm are referred respectively as the \emph{source} and \emph{meta}-embeddings.
ME has emerged as a promising ensemble approach to combine diverse pretrained source word embeddings to preserve their complementary strengths.

The problem settings of ME learning differ from that of source embedding learning in important ways as follows.
\paragraph{1. ME methods must be agnostic to the methods  used to train the source embeddings.\\}
The source embeddings used as the inputs to an ME learning method can be trained using different methods.
For example, context-insensitive \emph{static} word embedding methods~\citep{Dhillon:2015,Mnih:HLBL:NIPS:2008,Collobert:2011,Huang:ACL:2012,Milkov:2013,Pennington:EMNLP:2014} represent a word by a single vector that does not vary depending on the context in which the word occurs.
On the other hand, \emph{contextualised} word embedding methods~\citep{Elmo,BERT,XLNET,ALBERT,RoBERTa} represent the same word with different embeddings in its different contexts.
It is not clear beforehand which word embedding is best for a particular NLP task.
By being agnostic to the underlying differences in the source embedding learning methods, ME learning methods are in principle able to incorporate a wide range of source word embeddings.
Moreover, this decoupling of source embedding learning from the ME learning simplifies the latter.

\paragraph{2. ME methods must not assume access to the original training resources used to train the source embeddings.\\}
Source embeddings can be trained using different linguistic resources such as text corpora or dictionaries~\citep{tissier-gravier-habrard:2017:EMNLP2017,Alsuhaibani:AKBC:2019,Bollegala:AAAI:2016}.
Although pretrained word embeddings are often publicly released and are free of charge to use, the resources on which those embeddings were originally trained might not be publicly available due to copyright and licensing restrictions. 
Consequently, ME methods have not assumed the access to the original training resources that were used to train the source embeddings.
Therefore, an ME method must obtain all semantic information of words directly from the source embeddings.

\paragraph{3. ME methods must be able to handle pretrained word embeddings of different dimensionalities.\\}
Because ME methods operate directly on pretrained source word embeddings without retraining them on linguistic resources, the dimensionalities of the source word embeddings are often different.
Prior work~\citep{Yin:2018,Levy:TACL:2015} studying word embeddings have shown that the performance of a static word embedding is directly influenced by its dimensionality.
ME learning methods use different techniques such as concatenation~\citep{Yin:ACL:2016}, orthogonal projections~\citep{jawanpuria-etal-2020-learning,He:2020} and averaging~\citep{Coates:NAACL:2018} after applying zero-padding to the sources with smaller dimensionalities as necessary to handle source embeddings with different dimensionalities.

\paragraph{Applications of ME:}
ME learning is attractive from an NLP practitioners point for several reasons.
First, as mentioned above, there is already a large number of pretrained and publicly available repositories of static and contextialised word embeddings.
However, it is not readily obvious what is the best word embedding method to represent the input in a particular NLP application.
We might not be able to try each and every source embedding due to time or computational constraints.
ME learning provides a convenient alternative to selecting the single best word embedding, where we can use a ME trained from \emph{all} available source embeddings.
Second, \emph{unsupervised ME learning methods} (\autoref{sec:unsup}) do \emph{not} require labelled data when creating an ME from a given set of source word embeddings.
This is particularly attractive in scenarios where we do not have sufficiently large training resources for learning word embeddings from scratch but have access to multiple pretrained word embeddings.
Moreover, by using multiple source embeddings we might be able to overcome the limited vocabulary coverage in the individual sources.
Third, in situations where there is some labelled data for the target task or domain, we can use \emph{supervised ME learning methods} (\autoref{sec:sup}) to fine-tune the MEs for the target task or domain.

From a theoretical point-of-view, ME learning can be seen as an instance of ensemble learning~\citep{EnsembleLearning,Polikar_2012}, where we incorporate information from multiple models of lexical (word-level) semantics to learn an improved representation model.
An ensemble typically helps to cancel out noise in individual models, while reinforcing the useful patterns repeated in multiple models~\citep{muromagi-etal-2017-linear}. 
Although there are some theoretical work studying word embedding learning~\citep{Arora:TACL:2016,All-but-Top,Bollegala:AAAI:2018}, the theoretical analysis of ME learning has been under-developed, with the exception of concatenated meta-embeddings~\citep{Bollegala:IJCAIa:2022}.
For example, under what conditions can we learn a better ME than individual source embeddings is an important theoretical consideration.
ME learning can also been seen as related to \emph{model distillation}~\citep{46642,Hinton:2015} where we must learn a simpler \emph{student} model from a more complicated \emph{teacher} model.
Model distillation is an actively researched topic in deep learning where it is attractive to learn smaller networks involving a lesser number of parameters from a larger network to avoid overfitting and inference-time efficiency.

In this survey paper we focus on word-level ME learning.
We first define the ME problem (\autoref{sec:definition}) and cover unsupervised (\autoref{sec:unsup}) and supervised (\autoref{sec:sup}) ME methods.
We also look at multilingual ME in \autoref{sec:multi}.
Finally, we discuss the performance of different ME methods (\autoref{sec:eval}) and present potential future research directions (\autoref{sec:issues}).
Moreover, we publicly release a ME framework\footnote{\url{https://github.com/Bollegala/Meta-Embedding-Framework}} that implements several ME methods covered in this paper, which we believe would be useful to further enhance the readers' understanding on this emerging topic.

\section{Meta-Embedding -- Problem Definition}
\label{sec:definition}

Let us consider a set of $N$ source word embeddings $s_{1}, s_{2}, \ldots, s_{N}$ respectively covering vocabularies (i.e. sets of words) $\cV_1, \cV_2, \ldots, \cV_n$.
The embedding of a word $w$ in $s_{j}$ is denoted by $\vec{s}_{j}(w) \in \R^{d_{j}}$, where $d_{j}$ is the dimensionality of $s_{j}$.
We can represent $s_{j}$ by an embedding matrix $\mat{E}_{j} \in \R^{d_{j} \times |\cV_j|}$.
For example, $\mat{E}_{1}$ could be the embedding matrix obtained by running skip-gram with negative sampling~\citep[(SGNS)][]{Milkov:2013} on a corpus, whereas $\mat{E}_{2}$ could be that obtained from global vector prediction~\cite[(GloVe)][]{Pennington:EMNLP:2014} etc.
Then, the problem of ME  can be stated as -- \emph{what is the optimal way to combine $\mat{E}_{1}, \ldots, \mat{E}_{n}$ such that some goodness measure defined for the accuracy of the semantic representation for the words is maximised?}

The source word embeddings in general do not have to cover the same set of words.
If $w \notin \cV_n$, we can either assign a zero embedding or a random embedding as $\vec{s}_n(w)$ as a workaround. 
\cite{Yin:ACL:2016}  proposed a method to predict source embeddings for the words missing in a particular source.
Specifically, given two different sources $s_n$ and $s_m$ (where $n \neq m$) they learn a projection matrix $\mat{A} \in \R^{d_m \times d_n}$ using the words in $\cV_n \cap \cV_m$, the intersection between the vocabularies covered by both $s_n$ and $s_m$.
They find $\mat{A}$ by minimising the sum of squared loss, $\sum_{w \in \cV_n \cap \cV_m} \norm{\mat{A}\vec{s}_n(w) - \vec{s}_m(w)}_2^2$.
Finally, we can predict the source embedding for a word $w' \notin \cV_m$ and $w' \in \cV_n$ using the learnt $\mat{A}$ as $\mat{A}\vec{s}_n(w')$.
If we have multiple sources, we can learn such projection matrices between each pair of sources and in both directions.
We can then, for example, consider the average of all predicted embeddings for a word as its source embedding in a particular source.
After this preprocessing step, all words will be covered by all source embeddings.
\cite{Yin:ACL:2016} showed that by applying this preprocessing step prior to learning MEs (referred to as the 1\texttt{TO}N+ method in their paper) to significantly improve the performance of the learnt MEs.
However, as we see later, much prior work in ME learning do assume a common vocabulary over all source embeddings for simplicity.
Without loss of generality, we will assume that all words are covered by a common vocabulary $\cV$ after applying any one of the above-mentioned methods.

\section{Unsupervised Meta-Embedding Learning}
\label{sec:unsup}

In unsupervised ME  we do not assume the availability of any manually-annotated labelled data that we can use in the learning process.
In this setting all data that we have at our disposal is limited to the pretrained source embeddings.

\subsection{Concatenation}
\label{sec:conc}

One of the simplest approaches to create an ME under the unsupervised setting is vector concatenation~\citep{Bao:COLING:2018,Yin:ACL:2016,Bollegala:IJCAI:2018}.
Denoting concatenation by $\oplus$, we can express the concatenated ME, $\vec{m}_{\mathrm{conc}}(w) \in \R^{d_1 + \ldots + d_N}$, of a word $w \in \cV$ by \eqref{eq:conc}.
\begin{align}
\label{eq:conc}
\vec{m}_{\mathrm{conc}}(w) &= \vec{s}_{1}(w)\oplus  \ldots \vec{s}_{N}(w) \nonumber \\
& = \oplus_{j=1}^{N} \vec{s}_{j}(w)
\end{align}
\cite{AAAI:2016:Goikoetxea} showed that the concatenation of word embeddings learnt separately from a corpus and WordNet to produce superior word embeddings.
However, one disadvantage of using concatenation to produce MEs is that it increases the dimensionality of the ME space, which is the sum of the dimensionalities of the sources.
\cite{Yin:ACL:2016} post-processed the MEs created by concatenating the source embeddings using Singular Value Decomposition (SVD) to reduce the dimensionality.
However, applying SVD often results in degradation of accuracy in the MEs compared to the original concatenated version~\citep{Bollegala:IJCAI:2018}.

It is easier to see that concatenation does not remove any information that is already covered by the source embeddings.
However, it is not obvious under what conditions concatenation could produce an ME that is superior to the input source embeddings.
\cite{Bollegala:IJCAIa:2022} shows that concatenation minimises the pairwise inner-product \cite[(PIP)][]{Yin:2018} loss between the source embeddings and an idealised ME.
PIP loss has been shown to be directly related to the dimensionality of a word embedding, and has been used as a criterion for selecting the optimal dimensionality for static word embeddings.
They propose a weighted variant of concatenation where the dimensions of each source is linearly weighted prior to concatenation.
The weight parameters can be learnt in an unsupervised manner by minimising the empirical PIP loss.

\subsection{Averaging}
\label{sec:avg}

Note that as we already mentioned in \autoref{sec:intro}, source embeddings are trained independently and can have different dimensionalities.
Even when the dimensionalities do agree, vectors that lie in different vector spaces cannot be readily averaged.
However, rather surprisingly,  \cite{Coates:NAACL:2018} showed that accurate MEs can be produced by first zero-padding source embeddings as necessary to bring them to a common dimensionality and then by averaging to create $\vec{m}_{\mathrm{avg}}(w)$ as given by \eqref{eq:avg}.
\begin{align}
\label{eq:avg}
\vec{m}_{\mathrm{avg}}(w) = \frac{1}{N} \sum_{j=1}^{N} \vec{s}^{*}_j(w)
\end{align}
Here, $\vec{s}^{*}_j(w)$ is the zero-padded version of $\vec{s}_j(w)$ such that its dimensionality is equal to $\max(d_1, \ldots, d_N)$.
In contrast to concatenation, averaging has the desirable property that the dimensionality of the ME is upper-bounded by $\max(d_1, \ldots, d_N) < \sum_{j=1}^{N} d_j$.
\cite{Coates:NAACL:2018} showed that when word embeddings in each source are approximately orthogonal, a condition that they empirically validate for pre-trained word embeddings, averaging can approximate the MEs created by concatenating. 

Although averaging does not increase the dimensionality of the ME space as with concatenation, it does not consistently outperform concatenation, especially when the orthogonality condition does not hold.
To overcome this problem, \cite{jawanpuria-etal-2020-learning} proposed to first learn orthogonal projection matrices for each source embedding space.
They measure the Mahalanobis metric between the projected source embeddings, which is a generalisation of the inner-product that does not assume the dimensions in the vector space to be uncorrelated.

To explain their proposal further, let us consider two sources $s_1$ and $s_2$ with identical dimensionality $d$. 
Let us assume the orthogonal projection matrices for $s_1$ and $s_2$ to be respectively $\mat{A}_1 \in \R^{d \times d}$ and $\mat{A}_2 \in \R^{d \times d}$.
The two words $w_i, w_j \in \cV_1 \cap \cV_2$ are projected to a common space respectively as $\mat{A}_1\vec{s}_1(w_i) \in \R^d$ and $\mat{A}_2\vec{s}_2(w_j) \in \R^d$. 
The similarity in this projected space is computed using a Mahalanobis metric $(\mat{A}_1\vec{s}_1(w_i))\T \mat{B}  (\mat{A}_2\vec{s}_2(w_j))$ defined by the matrix $\mat{B} \in \R^{d \times d}$.
They learn $\mat{A}_1, \mat{A}_2$ and $\mat{B}$ such that the above metric computed between the projected source embeddings of the same word is close 1, while that for two different words is close to 0.
Their training objective can be written concisely as in \eqref{eq:geometric} using the embedding matrices $\mat{E}_1, \mat{E}_2 \in \R^{d \times |\cV_1 \cap \cV_2|}$ and a matrix $\mat{Y}$ where the $(i,j)$ element $Y_{ij} = 1$ if $w_i = w_j$ and $0$ otherwise.
\begin{align}
\label{eq:geometric}
\minimise_{\mat{A}_1, \mat{A}_2, \mat{B}} \norm{\mat{E}_1\T\mat{A}_1\T\mat{B}\mat{A}_2\mat{E}_2 - \mat{Y}}_F^2 + \lambda \norm{\mat{B}}_F^2
\end{align}
Here, $\lambda \geq 0$ is a regularisation coefficient corresponding to the Frobenius norm regularisation of $\mat{B}$, which prefers smaller Mahalanobis matrices.
They show that the averaging of the projected source embeddings (i.e. $(\mat{B}^{\frac{1}{2}} \mat{A}_1\vec{s}_1(w) + \mat{B}^{\frac{1}{2}}\mat{A}_2\vec{s}_2(w))/2$) to outperform simple non-projected averaging (given by \eqref{eq:avg}). 
Learning such orthogonal projections for the sources has shown to be useful even in supervised ME learning~\citep{He:2020} as discussed later in \autoref{sec:sup}.

\subsection{Linear Projections}
\label{sec:proj}

In their pioneering work on ME, \cite{Yin:ACL:2016} proposed to project source embeddings to a common space via source-specific linear transformations, which they refer to as 1\texttt{TO}N.
They require that the ME of a word $w$, $\vec{m}_{\text{1\texttt{TO}N}}(w) \in \R^{d_{m}}$, reconstruct each source embedding, $\vec{s}_j(w)$ of $w$ using a linear projection matrix, $\mat{A}_j \in \R^{d_j \times d_m}$, from $s_j$ to the ME space by as given by \eqref{eq:1ton}.
\begin{align}
\label{eq:1ton}
\hat{\vec{s}}_j(w) = \mat{A}_j \vec{m}_{\text{1\texttt{TO}N}}(w)
\end{align}
Here, $\hat{\vec{s}}_j(w)$ is the reconstructed source embedding of $w$ from the ME.
Next, the squared Euclidean distance between the source- and MEs is minimised over all words in the intersection of the source vocabularies, subjected to Frobenius norm regularisation as in \eqref{eq:loss-1ton}.
{\small
\begin{align}
\label{eq:loss-1ton}
\minimise_{\substack{\forall_{j=1}^{N} \mat{A}_j \\  \forall_{w \in \cV}\ \vec{m}_{\text{1\texttt{TO}N}}(w)}} \sum_{j=1}^{N} \alpha_j \left( \sum_{w \in \cV} \norm{\hat{\vec{s}}_j(w)  - \vec{s}_j(w)}_2^2 + \norm{\mat{A}_j}_{F}^{2} \right)
\end{align}
}%
They use different weighting coefficients $\alpha_j$ to account for the differences in accuracies of the sources.
They determine $\alpha_j$ using the Pearson correlation coefficients computed between the human similarity ratings and cosine similarity computed using the each source embedding between word pairs on the \cite{MCdataset} dataset.
The parameters can be learnt using stochastic gradient descent, alternating between projection matrices and MEs.

\cite{muromagi-etal-2017-linear} showed that by requiring the projection matrices to be orthogonal (corresponding to the Orthogonal Procrustes Problem) the accuracy of the learnt MEs is further improved.
However,  1\texttt{TO}N requires all words to be represented in all sources.
To overcome this limitation they predict the source embedding for missing words as described in \autoref{sec:definition}.

Assuming that a single \emph{global} linear projection can be learnt between the ME space and each source embedding as done by \cite{Yin:ACL:2016} is a stronger requirement. 
\cite{Bollegala:IJCAI:2018} relaxed this requirement by learning \emph{locally linear} (LLE) MEs.
To explain this method further let us consider computing the LLE-based ME, $\vec{m}_{\mathrm{LLE}}(w)$, of a word $w \in \cV_1 \cap \cV_2$ using two sources $s_1$ and $s_2$.
First, they compute the set of nearest neighbours, $\cN_j(w)$, of $w$ in $s_j$ and represent $w$ as the linearly-weighted combination of its neighbours by a matrix $\mat{A}$ by minimisng \eqref{eq:LLE-1}.
\par\nobreak
{\small
\begin{align}
\vspace{-5mm}
\label{eq:LLE-1}
 \minimise_{\mat{A}} \sum_{j=1}^{2} \sum_{w \in \cV_1 \cap \cV_2} \norm{\vec{s}_j(w) - \sum_{w' \in \cN_j(w)} A_{ww'} \vec{s}_j(w)}_2^2
\end{align}
}%
They use AdaGrad to find the optimal $\mat{A}$.
Next, MEs are learnt by minimising \eqref{eq:LLE-2} using the learnt neighbourhood reconstruction weights in $\mat{A}$ are  preserved in a vector space common to all source embeddings. 
\par\nobreak
{\small
\begin{align}
\vspace{-5mm}
\label{eq:LLE-2}
\sum_{w \in \cV_1 \cap \cV_2} \norm{\vec{m}_{\mathrm{LLE}}(w) - \sum_{j=1}^{2}\sum_{w' \in \cN_j(w)} C_{ww'}\vec{m}_{\mathrm{LLE}}(w')}_2^2
\end{align}
}%
Here, $C_{ww'} = A_{ww'} \sum_{j=1}^{2} \mathbb{I}[w' \in \cN_j(w)]$, where $\mathbb{I}$ is the indicator function which returns 1 if the statement evaluated is True.
Optimal MEs can then be found by solving an eigendecomposition of the matrix $(\mat{I} - \mat{C})\T(\mat{I} - \mat{C})$, where $\mat{C}$ is the matrix formed by arranging $C_{ww'}$ as the $(w,w')$ element.
This approach has the advantage that it does not require all words to be represented by all sources, thereby obviating the need to predict missing source embeddings prior to ME.

\subsection{Autoencoding}
\label{sec:ae}

\cite{Bao:COLING:2018} modelled ME learning as an \emph{autoencoding} problem where information embedded in different sources are integrated at different levels to propose three types of MEs: Decoupled Autoencoded ME (DAEME) (independently encode each source and  concatenate), Concatenated Autoencoded ME (CAEME) (independently decode MEs to reconstruct each source), and Averaged Autoencoded ME (AAEME) (similar to DAEME but instead of concatenation uses averaging).
Given the space constraints we describe only the AAEME model, which was the overall best performing among the three.

Consider two sources $s_1$ and $s_2$, which are encoded respectively by two encoders $E_1$ and $E_2$.
The AAEME of $w$ is computed as the $\ell_2$ normalised average of the encoded source embeddings as given by \eqref{eq:AAEME}.

\begin{align}
\label{eq:AAEME}
\vec{m}_{\mathrm{AAEME}}(w) = \frac{E_1(\vec{s}_1(w)) + E_2(\vec{s}_2(w))}{\norm{E_1(\vec{s}_1(w)) + E_2(\vec{s}_2(w))}_2}
\end{align}

Two independent decoders, $D_1$ and $D_2$, are trained to reconstruct the two sources from the ME.
$E_1, E_2, D_1$ and $D_2$ are jointly learnt to minimise the weighted reconstruction loss given by \eqref{eq:AE-loss}.
\par\nobreak
{\small
\begin{align}
\label{eq:AE-loss}
\minimise_{E_1,E_2,D_1,D_2} \sum_{w \in \cV_1 \cap \cV_2} \Large( & \lambda_1 \norm{\vec{s}_1(w) - D_1(E_1(\vec{s}_1(w)))}_2^2 + \nonumber \\
								& \lambda_2 \norm{\vec{s}_2(w) - D_2(E_2(\vec{s}_2(w)))}_2^2 \Large)
\end{align} 
}%
The weighting coefficients $\lambda_1$ and $\lambda_2$ can be used to assign different emphasis to reconstructing the two sources and are tuned using a validation dataset.
In comparison to methods that learn globally or locally linear transformations~\citep{Bollegala:IJCAI:2018,Yin:ACL:2016}, autoencoders learn nonlinear transformations. Their proposed autoencoder variants outperform 1\texttt{TO}N and 1\texttt{TO}N+ on multiple benchmark tasks.

Although our focus in this survey is word-level ME learning, sentence-level ME methods have also been proposed~\citep{poerner-etal-2020-sentence,Takahashi:LREC:2022}.
\citet{poerner-etal-2020-sentence} proposed several methods to combine sentence-embeddings from pretrained encoders such as by concatenating and averaging the individual sentence embeddings.These methods correspond to using sentence embeddings instead of source word embeddings in \eqref{eq:conc} and \eqref{eq:avg} with $\ell_2$ normalised sources.
Moreover, they used the Generalised Canonical Correlation Analysis (GCCA), which extends Canonical Correlation Analysis to more than three random vectors, to learn sentence-level MEs.
They also extend AAEME method described in \autoref{sec:ae} to multiple sentence encoders, where they learn an autoencoder between each pair of sources.
They found that GCCA to perform best in sentence similarity prediction tasks.
\citet{Takahashi:LREC:2022} proposed an unsupervised sentence-level ME method, which learns attention weights and transformation matrices over contextualised embeddings such that multiple word- and sentence-level co-occurrence criteria are simultaneously satisfied.

\section{Supervised Meta-Embedding Learning}
\label{sec:sup}

MEs have also been learned specifically for a set of supervised tasks. Unlike unsupervised ME learning methods described in \autoref{sec:unsup}, supervised MEs use end-to-end learning and fine-tune the MEs specifically for the downstream tasks of interest. 

\cite{neill2018meta} used MEs to regularise a supervised learner by reconstructing the ensemble set of pretrained embeddings as an auxiliary task to the main supervised task whereby the encoder is shared between both tasks.~\eqref{eq:recon_loss} shows the auxiliary reconstruction mean squared error loss weighted by $\alpha$ for each word $w_t$ in a sequence of of length $T$ words, and the main sequence classification task cross-entropy loss, weighted by $\beta$. Here,  $f_{\theta_{\mathrm{aux}}}$ is the subnetwork of $f_{\theta}$ that corresponds to the ME reconstruction and $f_{\theta_{\mathrm{main}}}$ is the subnetwork used to learn on the main task, i.e $f_{\theta}$ except the decoder layer of the ME autoencoder.
\begin{align}\label{eq:recon_loss}
\begin{gathered}
\minimise_{\mathbf{\theta}} \frac{1}{TN}\sum_{t=1}^{T}\Big[ \alpha \big( \mathbf{f}_{\theta_{\mathrm{aux}}}(\vec{m}_{\mathrm{conc}}(w_t)) - \\ \vec{m}_{\mathrm{conc}}(w_t)\big)^2 + \beta \sum_{c=1}^{C}\mathbf{f}_{\theta_{\mathrm{main}}}\big(\vec{m}_{\mathrm{conc}}(w_t))\log y_{t,c}\big) \Big]
\end{gathered}
\end{align}
The ME reconstruction shows improved performance on both intrinsic tasks (word similarity and relatedness) and extrinsic tasks (named entity recognition, part of speech tagging and sentiment analysis). 
They also show that MEs require less labelled data for Universal Part of Speech tagging\footnote{\url{https://universaldependencies.org/u/pos/all.html}} to perform as well as unsupervised MEs. 
\cite{wu2020task} also successfully deploy the aforementioned ME regularisation in the supervised learning setting.
They showed that as the ME hidden layer becomes smaller, the improvement in performance for supervised MEs over unsupervised MEs becomes larger. 

\cite{kiela-etal-2018-dynamic} proposed a supervised ME learning method where they compute the ME of a word using a dynamically-weighted sum of the projected source word embeddings.
Each source embedding, $s_j$, is projected to a common $d$-dimensional space using a matrix $\mat{P}_j \in \R^{d_j \times d}$ and a basis vector $\vec{b}_j \in \R^d$ as given by \eqref{eq:proj-dme}.
\begin{align}
\label{eq:proj-dme}
\vec{s}'_j(w) = \mat{P}_j\vec{s}_j(w) + \vec{b}_j
\end{align}
Next, the Dynamic Meta Embedding, $\vec{m}_{\mathrm{dme}}(w)$, of a word $w$ is computed as in \eqref{eq:dme}.
\begin{align}
\label{eq:dme}
\vec{m}_{\mathrm{dme}}(w) = \sum_{j=1}^{N} \alpha_{w,j}\vec{s}'_j(w)
\end{align}
Here, $\alpha_{w,j}$ is the scalar weight associated with $w$ in source $s_j$, and is computed via self-attention mechanism as given by \eqref{eq:attention}.
\begin{align}
\label{eq:attention}
\alpha_{w,j} = \phi(\vec{a}\T\vec{s}'_j(w) + b)
\end{align}
Here, $\phi$ is an activation function such as softmax, $\vec{a} \in \R^d$ and $b \in R$ are respectively source and word independent attention parameters to be learnt using labelled data.
They also proposed a contextualised version of DME (CDME), where they used the hidden state from a BiLSM that takes the projected source embeddings as the input.
Although the differences between DME and CDME were not statistically significant, overall, the highest maximum performances were reported with CDME.

\cite{xie2019dynamic} extended this approach by introducing task-specific factors for a downstream task by computing the pairwise interactions between embeddings in the ensemble set. 
\cite{He:2020} also create task-specific MEs by learning orthogonal transformations to linearly combine the source embeddings. 
As already discussed in \autoref{sec:avg}, enforcing orthogonality on the transformation matrix has shown to improve performance of ME.

\cite{FAME} proposed feature-based adversarial meta-embeddings (FAME) for creating MEs from both word- and subword-level source embeddings.
Specifically, a given token is represented by source embeddings as well as source-independent features related to the token such as frequency, length, shape (e.g. upper/lowercasing, punctuation, number etc.). 
Source embeddings are projected to a common vector space using linear projections, and their attention-weighted sum is computed as the ME.
A gradient reversal layer~\citep{DANN} is used to learn the projection and attention related parameters.
FAME achieves SoTA results for part-of-speech (POS) tagging in 27 languages.

\section{Multi-lingual Meta-Embedding Learning}
\label{sec:multi}

MEs have also been extended to the cross-lingual and multi-lingual settings.~\citet{winata2019learning} use  self-attention across different embeddings to learn  multi-lingual MEs. 
For Named Entity Recognition (NER), the multi-lingual embeddings in the ensemble set are concatenated and passed as the input into a transformer encoder and a conditional random field is used as the classification layer. 
The use of self-attention weights showed to improve over a linear layer without self-attention weights.~\citet{winata2019hierarchical} have also extended this to using hierarchical MEs (HMEs), which refers to word-level MEs that are created via a weighted average of sub-word level ME representations. They find that HMEs outperform regular MEs on code-switching NER through the use of pretrained subword embeddings given by $\mathtt{fasttext}$~\citep{FastText}. 

\citet{garcia2020common} learn MEs for cross-lingual tasks by projecting the embeddings into a common semantic space. Embeddings of resource-rich languages can then be used to improve the quality of learned embeddings of low-resourced languages. This is achieved in three steps: (1) align the vector spaces of different vocabularies of each language using the bilingual mapper $\mathtt{VecMap}$~\citep{artetxe-labaka-agirre:2016:EMNLP2016}, (2) create new embeddings for missing words in the source embeddings and (3) average the embeddings in the ensemble set. The resulting ME vocabulary will then be the union of the vocabularies of the word embeddings used. This method is referred to as MVM (Meta-VecMap).

\citet{doval2018improving} align embeddings into a bilingual vector space using $\mathtt{VecMap}$ and $\mathtt{MUSE}$~\citep{conneau2017word} and use a linear layer to transform the aligned embeddings such that the average word vector in the ensemble set can be predicted in the target language from the source. 
This is motivated by the finding that the average of word embeddings is a good approximation for MEs~\citep{Coates:NAACL:2018} as already discussed in \autoref{sec:avg}.

\section{Discussion}

In this section, we first discuss the performance of the different ME methods described in this paper.
Next, we discuss the limitations in existing ME learning methods and highlight potential future research directions.

\subsection{Evaluation and Performance}
\label{sec:eval}

Given that ME are representing the meaning of words using vectors, MEs have been evaluated following the same benchmark tasks commonly used for evaluating source (static) word embeddings such as word or sentence similarity measurement, analogy detection, text classification, textual entailment recognition, part-of-speech tagging~\citep{FAME}, etc.

The performance of an ME created using a set of source embeddings depends on several factors such as the source embeddings used (e.g, the resources used to train the sources, their dimensionalities),  dimensionality of the ME, and hyperparameters (and how they were tuned) for the downstream tasks (in the case of supervised ME).
However, prior work in ME learning have used different settings, which makes it difficult to make direct comparisons among results reported in different papers~\citep{garcia-ferrero-etal-2021-benchmarking-meta}.

Consistently across published results, concatenation has shown to be a competitive baseline, and averaging does not always outperform concatenation.
Predicting source embeddings for out-of-vocabulary words has reported mixed results~\citep{garcia-ferrero-etal-2021-benchmarking-meta}.
Methods for predicting source embeddings for missing embeddings in a source uses simple linear transformations such as learning projection matrices~\citep{Yin:ACL:2016,garcia-ferrero-etal-2021-benchmarking-meta}.
However, whether such transformations always exist between independently trained vector spaces is unclear~\citep{Bollegala:PLoS:2017}.
On the other hand, averaging has reported good performance without requiring the prediction of missing source embeddings because average is already a good approximation for the missing source embeddings.
However, scaling each dimension of a source embedding to zero mean and subsequently normalising the embeddings to unit $\ell_2$ norm is required when the dimensionalities and norms of the source embeddings that are averaged are significantly different. 

Moreover, carefully weighting sources using some validation data (e.g. performances obtained on a benchmark dataset with individual sources) has shown to improve performance of concatenation~\citep{Yin:ACL:2016}.
Although applying SVD reduces the dimensionality of the concatenated MEs, it does not always outperform the concatenation baseline.
In particular, the number of singular values remains an important factor that influences the performance of this method~\citep{Bollegala:IJCAIa:2022}.
The best performance for unsupervised ME has been reported by autoencoding methods, and in particular by AEME~\citep{Bao:COLING:2018}.
Overall, supervised or task-specific ME learning methods have reported superior performances over unsupervised ones~\citep{FAME} in tasks such as sentence classification, POS tagging and NER.
Therefore, when there is some labelled data available for the target task, we recommend using supervised ME methods.
However, it remains unclear whether a supervised ME trained for one particular task will still perform well for a different task.

\subsection{Issues and Future Directions}
\label{sec:issues}

ME learning is still a growing and active research area. 
Next, we outline existing issues and suggest potential research directions for future work on this topics.

\paragraph{Contextualised MEs:}
Despite the good performance on intrinsic evaluation tasks,
\citet{garcia-ferrero-etal-2021-benchmarking-meta} showed that none of the ME methods outperformed $\mathtt{fasttext}$ source embedding on GLUE benchmarks~\citep{GLUE}.
Moreover, concatenation (with centering and normalising of source embeddings) and averaging have turned out to be strong baselines, often outperforming more complex ME methods.
Given that contextualised embeddings obtain the SoTA performances on such extrinsic evaluations, we believe it is important for future research in ME learning to consider contextualised embeddings as sources instead of static word embeddings~\citep{Takahashi:LREC:2022,poerner-etal-2020-sentence}.

\paragraph{Sense-specific MEs:}
Thus far, ME learning methods have considered a word (or a sentence) as the unit of representation.
However, the meaning of a word can be ambiguous and it can have multiple senses. 
Sense-embedding learning methods learn multi-prototype embeddings~\citep{Reisinger:NAACL:2010,neelakantan-EtAl:2014:EMNLP2014} corresponding to the different senses of the same ambiguous word.
How to combine sense embeddings with word embeddings to create sense-specific MEs remains an open research problem.

\paragraph{MEs For Sequence-to-Sequence Models:}
MEs have yet to be used for sequence-to-sequence generation tasks such as machine translation
Therefore, we predict that a study of how they can be used in tandem with SoTA models such as the Transformer~\citep{vaswani2017attention} would be an impactful contribution. 
Particular questions of interests would be: 
(1) \textit{How do Transformers perform on text generation tasks when preinitialised with MEs}? and
(2) \textit{Does a smaller model preinitialised with MEs outperform a model without MEs}?.
Answering the aforementioned questions gives a clear indication of how MEs can be retrofitted into SoTA models and how they can obtain near SoTA results with shallower models.

\paragraph{Mitigating Negative Transfer:}
Creating an ME using all sources could lead to \emph{negative transfer}~\citep{Pan:TKDE:2010} and consequently degrade model performance. 
Although attention-based source-weighting methods have been proposed~\citep{xie2019dynamic,winata2019hierarchical},
that learn different weights for the sources, a systematic analysis of how negative transfer affects ME is required.

\paragraph{Theoretical Analysis of MEs:}
Compared to the numerous empirical studies, theoretical understanding of ME learning remains under-developed.
Some of the important theoretical questions are: (1) \textit{What is the optimal dimensionality for MEs to balance the memory vs performance tradeoff}?,
(2) \textit{What is the relative contribution between sources vs. ME learning algorithm towards the performance gains in downstream tasks}?,
 and (3) \textit{What are the generalisation bounds for the performance of ME learning algorithms beyond a specific set of sources}?.

\paragraph{Social bias in MEs:}
Word embeddings have shown to encode worrying levels of unfair social biases such as gender, racial and religious biases~\citep{kaneko-bollegala-2019-gender,Tolga:NIPS:2016}.
Given that MEs are incorporating multiple sources and further improve the accuracy of the embeddings, an unaddressed concern is whether an ME learning method would also amplify social biases contained in the source embeddings.
It would be ethically inappropriate to deploy MEs with social biases to downstream NLP applications.
We believe that further research is needed to detect and mitigate such biases in MEs.

\section{Conclusion}
\label{sec:conclusion}
We presented a survey of ME learning methods. 
We classified prior work into different categories such as unsupervised, supervised, sentence-level and multi-lingual ME learning methods.
Finally, we highlighted potential future research directions.
Given that ME learning is an active research topic, we hope this survey facilitates newcomers on this topic as well as providing inspiration to future developments in the broader AI community, incorporating existing word/text representations to create more accurate versions.

\bibliographystyle{named}
\bibliography{Embed}

\end{document}